\documentclass{llncs}

\input{psfig.sty}

\usepackage{url}

\begin{document}

\pagestyle{empty}
\mainmatter

\title{Anticipatory Guidance of Plot}

\author{Jarmo Laaksolahti \and Magnus Boman}

\institute{Swedish Institute of Computer Science\\
Box 1263, SE-164 29 Kista, Sweden\\
\email{\{jarmo,mab\}@sics.se}}

\maketitle

\begin{abstract}
An anticipatory system for guiding plot development in interactive narratives is described. The executable model is a finite automaton that provides the implemented system with a look-ahead. The identification of undesirable future states in the model is used to guide the player, in a transparent manner. In this way, too radical twists of the plot can be avoided. Since the player participates in the development of the plot, such guidance can have many forms, depending on the environment of the player, on the behavior of the other players, and on the means of player interaction. We present a design method for interactive narratives which produces designs suitable for the implementation of anticipatory mechanisms. Use of the method is illustrated by application to our interactive computer game Kaktus.
\end{abstract}

\section{Introduction}
\label{sect:internarr}
Interactive narrative is a form of entertainment that invites users to step into and interact with a fictive world. In contrast to traditional non-interactive narratives, participants in an interactive narrative are active in the creation of their own experiences. Through their actions, players interact with other agents (some of which might be artificial) and artifacts in the world---an experience that can be compared to role-playing or acting. Interactive narrative promises to empower players with a greater variety of offerings but also the capacity to deal with them \cite{vo00}, leading to deeper and more engaging experiences. 
 
Stories may be told in many different ways, depending on the order in which events are disclosed to the player. Plot control concerns itself with deciding which event of a story to present next, not to create entirely new stories (cf. \cite{winybo01}). The player should have the feeling that anything may happen while being nudged through the story along various story arcs \cite{ga95}. At the same time the player should feel that the choices she makes has a non-trivial impact on how the plot unfolds.

Plot guidance involves searching among a possibly huge amount of unfoldings for one that fulfills (some of) the author's intentions for the story~(cf. \cite{we97}). The search of the state space quickly becomes intractable, however, as the number of scenes grow. For a scenario consisting of as little as 16 scenes (with a normal movie having 40-60 scenes) search would have to consider billions of states. Limiting the depth of the search can reduce the size of the state space but can also result in bad plot unfoldings (i.e., the player can get stuck).

We hypothesize that anticipatory systems can provide an alternative means to efficient plot guidance. Robert Rosen suggests that the simplest way for anticipations to affect the properties of a dynamic system $S$ through a model $M$ is the following: \begin{quote}Let us imagine the state space of $S$ (and hence of $M$) to be partitioned into regions corresponding to ``desirable'' and ``undesirable'' states. As long as the trajectory in $M$ remains in a ``desirable'' region, no action is taken by $M$ through the effectors $E$. As soon as the $M$-trajectory moves into an ``undesirable'' region [...] the effector system is activated to change the dynamics of $S$ in such a way as to keep the $S$-trajectory out of the ``undesirable'' region. \cite{ro74} (p.247)\end{quote}

The system does not have to consider full trajectories but only those that start in the current state of $S$ and stretch some time into the future. This gives the system a look-ahead similar to that provided by search. By contrast, Rosen's description suggests that the main concern for an anticipatory system is to detect and avoid  bad trajectories instead of finding good ones. Hence, ideally the anticipatory system will not interfere with the object system's execution. 

We will, in Section~\ref{sect:kaktus}, present our own interactive narratives project, by means of an example. This example is then used as our object system $S$ in an anticipatory system for guiding players through interactive plot development. The details of an automaton model $M$ and effectors $E$ of the anticipatory system are presented in Section~\ref{sect:anti}. Since we present work in progress, we conclude by identifying topics for future research.

\section{A {K}aktus Scenario}
\label{sect:kaktus}
Our ongoing Kaktus project aims to create truly interactive socio-emotionally rich stories in which the player actively participates and influences the plot in non-trivial ways. We strive to move away from the simplistic ways of interacting found in many games, and instead focus on social and emotional interaction. 

The scenario centers around three teen-aged girls named Karin, Ebba, and Lovisa. The player of Kaktus acts as one of the girls while the system controls the others. We enter the story at a time when the girls are planning to organize a party for their friends. The story evolves over a period of several days prior to the party. During this period decisions have to be made concerning different matters related to the party, for instance, whom to invite, negotiating with parents about lending the house (or some other locale), what kind of food (if any) to serve, if alcohol should be allowed at the party, choice of music, etc. In addition to organizing the party, the game involves establishing and maintaining social relationships with the characters of the game. In order to be successful, the player must adopt the role of a teenage girl, be sensitive to the social and emotional cues in the environment, and act on the basis of those. 

In the traditional arts, a story is described as a sequence of events that through conflict changes dramatic values of importance for the characters of the story (see e.g., \cite{ch78,mc97}). Values are typically on a spectrum between counterpoints (e.g., love/hate, death/life, courage/cowardice, strength/weakness, good/evil). For example, the Kaktus scenario revolves around the values 
\begin{itemize}
	\item \textbf{love/hate} is illustrated by Lovisa's secret love for a boy in the local hockey team which may change due to events in the game.
	\item \textbf{friendship/enmity} is of great importance in the scenario. Plot unfolding---and ultimately success in arranging the party---depends on the players ability to interpret and manipulate the social configurations between characters.
	\item \textbf{boredom/exhilaration} is one of the main driving forces of the game. Boredom is the main reason for the girls to organize the party.
\end{itemize}

Story events are classified according to how much they change story values. Using the terminology of Robert McKee \cite{mc97}, the smallest value changing event is the \emph{beat}. It consists of an action/reaction pair that causes minor but significant changes to at least one story value. A \emph{scene} is built from beats and causes moderate changes to values. Scenes are the most fundamental building blocks of stories forming the arc of the story, or \begin{quote}the selection of events from characters' life stories that is composed into a strategic sequence to arouse specific emotions and to express a specific view of life. (\cite{mc97}, p.33)\end{quote} Next comes the scene \emph{sequence} and \emph{act}. Acts are strings of sequences causing major value reversals. Finally the \emph{story} itself is a long irreversible change of values. 

%Similarly the relevant parts of the story can be divided into seven scenes:
Using this definition of story events we can outline the story of our scenario through the following scenes:  
\begin{enumerate}
\item[$q_1$.] Introduction of Karin, Ebba and Lovisa, their current state of mind, and conception of the party idea.
\item[$q_2$.] Karin and Lovisa find out that Ebba cannot afford to organize a party.
\item[$q_3$.] Lovisa's secret love for Niklas is revealed to Karin.
\item[$q_4$.] Plans are made to hold the party at Lovisa's house.
\item[$q_5$.] How to get hold of alcohol to the party is discussed.
\item[$q_6$.] Karin, Ebba, and Lovisa invite people to the party.
\item[$q_7$.] The girls decide not to have a party.
\end{enumerate}

Each scene consists of beats detailing its content. Minimally a scene consists of beats detailing responses to input from the player. For instance, lines of dialog exchanged between characters and the player, or beats implementing (inter-) agent behaviors, such as behaviors for two agents to have a row. We will not go further into details regarding beats here.   

Some events in a story are more important than others. \emph{Kernels} \cite{ch78} or \emph{story-functional} events are crossroads where the plot of a story is forced to branch of in one direction or the other. By contrast \emph{satellites} or \emph{descriptive} events do not affect the plot in the same way. Instead, they serve as instruments for creating atmosphere, elaborating on kernels or fleshing out the story. In our scenario descriptive events are scenes where the user explores the narrative world she is immersed in, e.g., by chatting with the characters and getting to know them. Such events do not directly further the plot but the story would nevertheless be impoverished without them. 

\section{System Overview}
\label{sect:overview}
There is a fundamental conflict between interactivity and narrative. In narrative media experiences, such as suspense, comic effects or sympathy often depend on the ability to specify a sequence of events in the `right order'. In interactive narratives, in which the player occasionally takes control of what will happen next, such effects are difficult to achieve. A player, in a few mouse clicks, may ruin an experience that has painstakingly been designed by an author. By limiting the amount of options, a player may be pushed along an intended path, but at the same time such a design will decrease her amount of influence over the story.

Stories are ultimately about characters, and hence agents have long been considered natural elements in many story-telling systems \cite{ba92,boca01,ga95,havahu97,pamapr01}. To date there has been a fair amount of research regarding various aspects of interactive narratives. However, most agent-based interactive narrative systems have adopted a locally reactive approach, where individual agents within the system ground their actions in their own perception of the current situation \cite{lo97}. Comparatively little effort has gone into research on how to integrate local reactivity with a global deliberative plot control mechanism \cite{mast00,we97}. Selecting what events to recount and how to order them is what makes up the plot of a story. Kaktus adopts an agent-based approach to story-telling where characters in the guise of artificial agents or human players have a central role. Characters are the primary vehicles for conveying dramatic action and progress along different story arcs. As the scenario relies heavily on social and emotional interaction, artificial agents are required to simulate human behavior to some extent and be capable of `reasoning' about emotions, interpersonal relationships, and consequences of actions. 

\subsection{Playing the Game}
\label{sect:gameplay}
Game play in Kaktus is best described as a mix between role playing and simulation. It resembles role playing in that players will not act as themselves in the virtual world, but place themselves in the position of a fictive protagonist. The embodied agent will not be an instantiation of the user but rather a character. The player will act not so much as herself but as she thinks the character would act in any given situation. 

Kaktus is also inspired by simulation games in that artificial characters have a (simulated) life of their own including emotions, goals, plans, and social relations which are all affected by players' choices and actions, as in The SIMS (\url{http://www.thesims.com}). In fact, we envision that Kaktus may be played in different ways. Instead of having party arrangement as the primary goal of a game session, some players may instead choose to, e.g., maximize the friendship relations between characters. In this mode of game play Kaktus can act as a social simulator where players can---within a limited domain---hone their social skills. 

Players interact with the game mainly through dialog. Players type out text that is visible on screen while the artificial characters' dialog is synthesized speech. The discourse is real-time and players can use a number of objects available in the game (such as keys, diaries, candy, mobile phones, and buddy lists in those mobile phones) to accomplish tasks. 

%\subsection{Story and Anticipation}
%\label{sec:StoryAndAnticipation}

%Det här påverkas av story functional och non-story functional events. Det skall in någonstans.
%The consequence for plot guidance is that we do not have to over specify the users experience. We do not have to specify every desirable trajectory, although the main plot trajectories need to be. Nor do we need to specify every undesirable trajectory, although especially interesting or common ones may need to be~(see \ref{sect:topdowndesign}). Instead we can let players have freedom to discover their own path without the system constantly interfering but without compromising the quality of the story.  

%XXX more here? Anytime egenskaperna. 

%\begin{figure}
%\centerline{\psfig{figure=storyarch.eps,height=8.2cm}}
%\caption{The task of plot guidance}
%\label{fig:plotguidance}
%\end{figure}
 
\subsection{System Implementation}
\label{sec:implementation}

Kaktus uses a Belief Desire Intention (BDI) approach to model agents \cite{brispo91}. This approach gives agents a rudimentary personality in that they pursue subjective goals, based on equally subjective models of their environment. The core of our system is based on the JAM (for Java Agent Model) agent architecture \cite{hu99} to which we supply an extension for doing anticipatory planning. 

\begin{figure}
\centerline{\psfig{figure=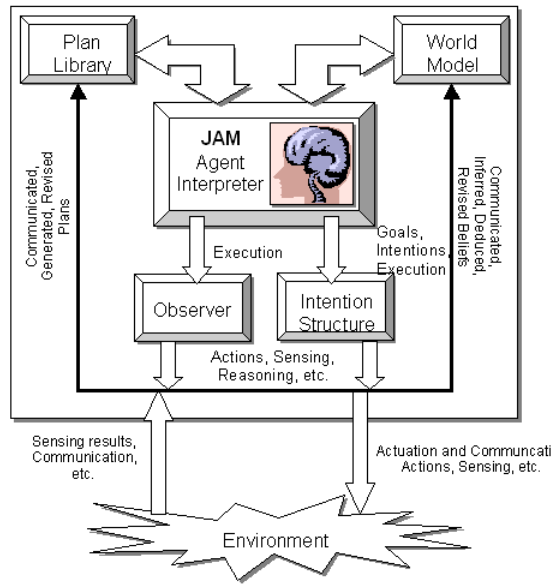,height=8.2cm}}
\caption{The anatomy of a JAM agent. Picture taken from \cite{hu01}.}
\label{fig:jamanatomy}
\end{figure}

The anatomy of a JAM agent is divided into five parts: \emph{a world model}, \emph{a plan library}, \emph{an interpreter}, \emph{an intention structure}, and \emph{an observer} (see Figure~\ref{fig:jamanatomy}). The world model is a database that represents the beliefs of the agent. The plan library is a collection of plans that the agent can use to achieve its goals. The interpreter reasons about what the agent should do and when it should do it. The intention structure is an internal model of the goals and activities the agent currently has committed itself to and keeps track of progress the agent has made toward accomplishing those goals. The observer is a special plan that the agent executes between plan steps in order to perform functionality outside of the scope of its normal goal/plan-based reasoning. 

\begin{figure}
\centerline{\psfig{figure=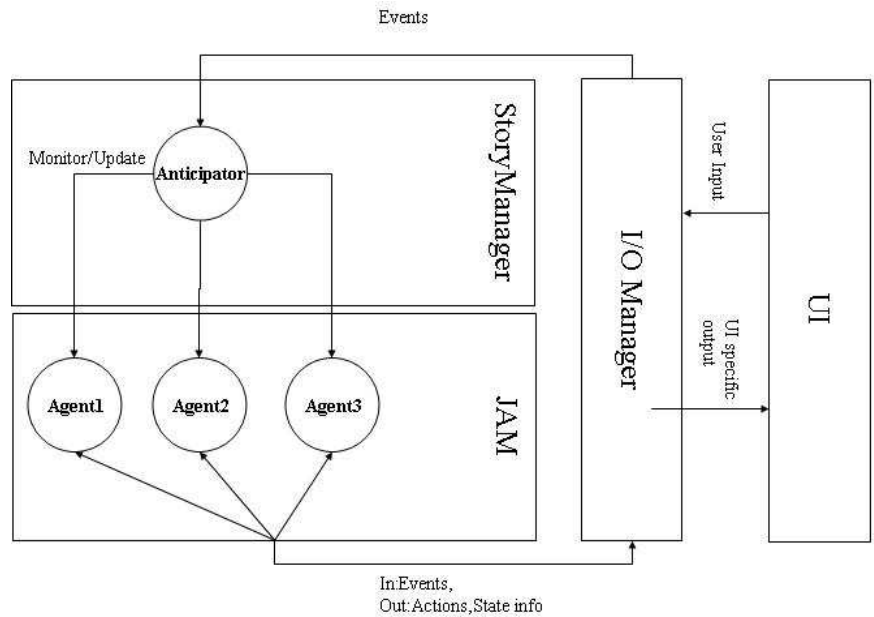,height=8.2cm}}
\caption{Overview of the system.}
\label{fig:sysdetails}
\end{figure}

A high level overview of the system is given in Figure~\ref{fig:sysdetails}. The system consists of four basic components:

\begin{itemize}
	\item A graphical front-end/user interface
	\item An I/O manager 
	\item A story manager
	\item An ensemble of JAM agents
\end{itemize}

The front-end is separated from the rest of the system according to a Model-View pattern. This facilitates having more than one type of interface to the system. Currently we are using an interface implemented in Macromedia Director but other interfaces, e.g.,  a web interface, are planned. Depending on the capabilities of the front-end some low-level functionality may be implemented in it, e.g., low-level movement primitives for agents.

The I/O Manager sits between the front-end and the story manager. The tasks performed by the I/O Manager vary depending on the type of front-end used. Minimally it consists of converting user input to internal formats and system output to front-end specific formats. For instance, text that the user has typed is converted to dialog moves and output is converted to formats such as Facial Animation Parameters (FAP), Body Animation Parameters (BAP) or ID's of animation files to play.

After conversion to an internal format user interface events are forwarded to the story manager. The story manager acts on the information it receives and continuously reflects the state of the system back to the front-end. Each character in the story is represented by a JAM agent--external to the story manager-- with a distinct set of goals, plans and intentions. During game play agents act autonomously, but are from time to time given direction by the story manager. 
%Note that there is only one anticipator; all the other (artificial) agents in the system are controlled by this single instance.

The story manager embodies the anticipatory planning mechanisms of the system through the \emph{anticipator} (cf. \cite{da03}). The anticipator monitors the progress of the story and decides on appropriate courses of action. The decisions are enforced by modifying the state of each agent---or tuning other system parameters---in order to accomplish the intended course of action. This includes, but is not limited to, adding or retracting information from an agent's world model, adding or deleting plans from an agent's plan repository and adding or deleting goals from an agent's list of goals. 

We utilize the observer functionality of JAM agents to communicate with the anticipator giving it a chance to inspect and modify the state of an agent between each planning step. While this scheme does not provide the anticipator with uninterrupted insight into an agents mind it is sufficient for our needs, since in effect the anticipator can make changes to the agents state \emph{before} the agent chooses and/or executes an action. 

Other distributions of responsibility are also possible, e.g., the agent community could manage without a super agent deciding on appropriate courses of action through voting or the super agent can act more as an advisor, guiding agents when they are uncertain of what to do (cf. \cite{bodakuve00}).

The anticipator uses copies of each agent to simulate execution of the system. The simulation has no side effects, such as output to the front-end. In an interactive application such as ours, generating output and waiting for input are operations that typically make up the bulk of execution time. In a simulation these operations can, and should, be left out, thus freeing time for running the simulation itself. Hence we do not anticipate a need for creating simpler models of our agents, in order to achieve faster than real time performance for the anticipator.
 
In addition to the anticipator, the story manager contains a model of desired plot unfoldings.  The plot model is encoded as a finite automaton which is described in detail in Section~\ref{sect:themodelm}. The model includes bookkeeping aspects of the system state such as which scenes have been played, which scenes are currently playable, and which scene is currently active. 

Each scene contains a set of beats encoded as JAM plans and goals. When a scene first becomes active agents in the scene are given goals and plans that enables them to perform the scene. Conversely, when a scene becomes inactive goals and plans specific to that scene are removed from agents' plan libraries and goal stacks. In this way the story manager portions out directions for the story in suitable chunks.

To illustrate how the anticipator works we will describe the sequence of events that takes place during execution of the system. 

%Based on the discussion above we define a narrative system containing three basic components: a set of story values where each value pertains to at least one character, a cast consisting of agents representing characters in the story, and a set of scenes. 
%Various bookkeeping aspects of the system state such as which scenes have been played, scenes that are playable, and which scene is currently active are all encoded in the finite automaton. For instance, a scene $q_i$ is playable if there is a transition leading from the current state to $q_i$ and all preconditions of $q_i$ are fulfilled.
\subsection{Example}
\label{sect:AnExample}
%The anticipator holds a model of each agent's state as well as a model of desired plot unfoldings. The plot model is encoded as a finite automaton which is described in detail in Section~\ref{sect:themodelm}. The model includes bookkeeping aspects of the system state such as which scenes have been played, which scenes are currently playable, and which scene is currently active. To illustrate how the anticipator works we will describe the sequence of events that takes place during execution of the system.

After the system is started each agents \emph{observer} plan will eventually be executed. The observer calls a synchronization method in the \emph{anticipator} passing a reference to the agent as a parameter. Through the reference the anticipator has access to the agent's world model, plan library, and intentional structure, which together control the behavior of the agents. 

The anticipator copies information from the agent, such as facts from its world model or the current goal of the agent, and stores it in a copy of the agent. The copy is later used for making predictions about the agents future behavior. If there are several agents in the system, the anticipator waits for all of them to call the synchronization method in order to get a snapshot of each agent's state, before relinquishing control.

Next, the anticipator starts a simulation of the system using the copied information. At regular intervals each agent observer plan  calls the synchronization method again. If the anticipator, based on it's simulations, predicts that the system will end up in an undesired state, it searches for an appropriate effector to apply (see Section~\ref{sect:theeffectorse}). In case there is more than one agent using the anticipator, it waits for all of them (or at least the ones that are affected by the chosen effector) to call the synchronization method in order to gain access to the entire system state, before it applies the effector. 

Given that there is a user involved, providing input which may be hard for the system to predict,  the synchronization can also act as a sensibility check of the anticipator's predictions. If the actual state does not evolve according to predictions the anticipator can discard the current predictions, gather new information, and start the cycle anew. 

\begin{figure}
	\begin{center}
	%\begin{minipage}{9cm}
	%\begin{footnotesize}
	\begin{verbatim}
GOALS:
  ACHIEVE live;
FACTS:
  FACT friends "Lovisa" "Karin" 1;
  FACT in_love "Lovisa" "Niklas";
PLAN:
{
NAME:
  "live"
GOAL:
  ACHIEVE live;
BODY:
  FACT friends "Lovisa" "Karin" $strength;
  OR
  {
    TEST( > $strength 1);
    ACHIEVE gossip;
  }
  {
    EXECUTE doIdle;
  };
}
PLAN:
{
NAME:
  "gossip"
GOAL:
  ACHIEVE gossip;
BODY:
  RETRIEVE in_love "Lovisa" $who;
  PERFORM tell "Karin" "in_love" "Lovisa" $who;
EFFECTS:
  ASSERT knows "Karin" "in_love" "Lovisa" $who;
}
\end{verbatim}
%\end{footnotesize}
%\end{minipage}
\end{center}
	\caption{An example of a JAM agent.}
	\label{fig:planexample}
\end{figure}

As an example consider the simple agent described in  Figure~\ref{fig:planexample}. This agent has the single top-level goal of \emph{ACHIEVE live} to pursue. The goal is achieved through displaying idle time behavior or, if the agent is on friendly terms with Karin, \emph{gossip} about Lovisa's infatuation with Niklas.

Let us suppose that in the anticipator's simulation this agent has come to a point where it has selected \emph{gossip} as its next plan to execute. However, suppose further that global story constraints dictate that for the time being this is not a good plan, since it leads to an undesirable state. The anticipator then starts a search for an effector (or effectors) that can prevent this situation from arising. In this case three effectors are found: lower the friendship value between the agent and Karin thus preventing the plan from becoming active, remove the plan from the agent's plan library (or replace it with a less harmful one) or give the agent a new goal with a higher priority. For simplicity let us suppose that the anticipator chooses to lower the friendship value between the agent and Karin. After this is done the agent resumes normal execution but with a new value on the friendship relation to Karin. At some later point in time when the agent would normally have selected the gossip plan for execution it now  displays idle time behavior instead, and the undesired state is avoided. This cycle is repeated until the user stops playing or the plot has reached its conclusion.

The same basic procedure is also applicable when more than one agent is involved. However it is likely that different effectors will need to be applied to each agent, resulting in a larger number of applied effectors than in the single agent case. For instance, to initiate a fight between two agents regarding some matter, they will need opposing views of the matter at hand, different knowledge, different arguments, etc. 

\section{The Anticipatory System}
\label{sect:anti}
After describing, in the classical Rosen sense (cf. \cite{ro85}), the model and the set of effectors for the object system $S$ (represented by the Kaktus scenario), we will concentrate on their significance to the design of interactive narratives.

Using the classification scheme introduced in \cite{busige02} we can regard our system as a \emph{state anticipatory} system. Through simulation, the anticipator forms explicit predictions about future states in order to avoid undesirable ones. Currently our model is limited in that we do not explicitly model players in any way. However, modeling users is notoriously difficult. Hence we will rely on empirical tests of the current system to indicate whether such an extension would be worth the added effort.
 
\subsection{The Model $M$}
\label{sect:themodelm}
\begin{figure}
\centerline{\psfig{figure=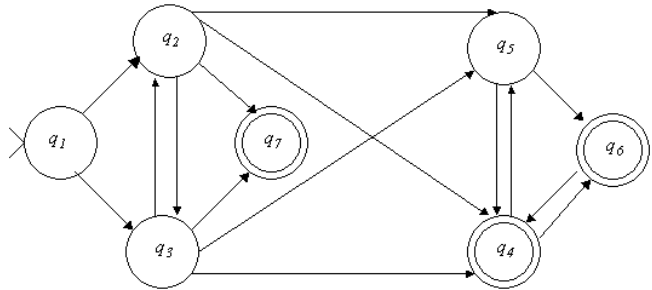,height=4.2cm}}
\caption{A state diagram for the finite automaton $M$, with only desirable states.}
\label{fig:finite1}
\end{figure}

Our model $M$ is a finite automaton, in which each state corresponds to a scene.  In Figure~\ref{fig:finite1}, the seven scenes of the Kaktus scenario are represented as states $q_1$ through $q_7$. Note that $M$ only contains story-functional scenes as described in section~\ref{sect:kaktus}. Descriptive events are accessible from most of the kernel events. However, since they do not influence the plot they are not included here. The start state is $q_1$, while the set of end states is \{$q_4,q_6,q_7$\}. The design method is top-down, in that $M$ was completed only after the key scenes had been identified. Note that one may also proceed bottom-up, letting $M$ depict sequential plot development from the start state to the end state of $M$. This entails that the design of $M$ is more important than design of the plot. In this case, the states of $M$ are compositional and scenes are identified as natural halts or crossroads in the evolution of the plot. Hence, if plot emergence is studied, a bottom-up design seems adequate. For conventional interactive game design, however, top-down is the default choice.

%Descriptive events are accessible from most of the kernel events. However, since they do not influence the plot they are not included here.

Finite automata also lend themselves to non-sequential development of plot. Models have previously been restricted to directed acyclic graphs \cite{we97}. We argue that instead it should be regarded as an opportunity. Many forms of interaction build their narrative framework on plot cycles, e.g., the repeated mouse chases in Tom \& Jerry cartoons. 

We will show below that problems related to computational complexity and model design complexity can be avoided. In fact, we have harsh constraints for the execution of $M$, since the time complexity must be low enough to allow for faster than real-time execution.

The state transitions in $M$ are labeled $a_0, a_1,\dots$ indicating that each transition is different. (For clarity of exposition, we have not labeled the edges in Figure~\ref{fig:finite1}.) In practice, there might be simple scenarios or small parts of an interactive drama in which transitions may be reused. However, for scenarios of the size and complexity we consider, uniqueness can be stipulated, especially since this stipulation has no mathematical significance. Each transition $a_m$ from $q_i$ to $q_j$ could be interpreted as (an ordinary Hoare triple with) pre- and postconditions. The precondition is the state described as $q_i$ plus the added conditions described in $a_m$, and the postcondition is simply $q_j$. 

Each transition $a_m$ corresponds to a set of conditions that can be either $true$ or $false$. Conditions are represented by one of the symbols \{$0$,$1$,$?$\} denoting $false$, $true$ and $either$ respectively. A set of conditions is represented by a fixed length string where each position in the string corresponds to a particular condition.  This means that transitions encode the desired value for all conditions in the system. However the wild card symbol '?' can be used in circumstances when the value of a condition is irrelevant to a transition. In our implementation conditions are represented by Java classes that are stored in a database. Each condition is evaluated at most once each cycle. The result of the evaluation is stored internally in the condition and can be reused later on. Currently we have implemented the following general purpose condition types:
\begin{itemize}
	\item Range 
	\item Boolean 
	\item Greater 
	\item Less
	\item Equal
\end{itemize}

These classes can be used to place conditions on any system parameters including story values, e.g., to prevent a transition from one climactic scene to another.

In addition we have implemented a set of conditions specific to agents that relate directly to facts, goals, plans, and emotions that an agent may have.
\begin{itemize}
	\item Knows
	\item Feels
	\item HasGoal
	\item HasPlan
\end{itemize}

The language that $M$ accepts is the set of words accepted. Each word is a sequence of names of transitions, e.g., $a_0a_3a_4a_3a_4a_6$. Therefore, an ordinary finite automaton suffices. Should any manipulation be required in $M$, a finite state transducer (or if the language was very complicated, a Turing machine) would be required (see, e.g., \cite{si99}). The preconditions described in $a_m$ are usually fairly complicated, and we will describe only one transition in the automaton pertaining to our Kaktus scenario. 

The transition we will describe takes us from a state where Karin (the player) does not 
know about Lovisa's romantic interest in Niklas to one where Ebba tells her about it. 
This is a climactic scene where many story values temporarily are at their extremes. 
There are several preconditions that must be fulfilled for this transition to take place. 
Assuming that the transition is named $a_m$ let us consider the set of preconditions 
$\{p(a_m)\}_j$. 
\begin{itemize}
\item $\{p(a_m)\}_1$ states that Karin must be unknowing about Lovisa's interest in Niklas, or the transition
will not make any sense.
\item $\{p(a_m)\}_2$ tells us that Lovisa must be unwilling to have the party, since getting to know about
and later inviting Niklas to the party is a way of persuading Lovisa.
\item $\{p(a_m)\}_3$ requires that Ebba actually wants to have the party. If she does not, she may not have 
a strong enough incentive to break her silence and tell Karin about Lovisa and Niklas.
\item $\{p(a_m)\}_4$ states that Ebba must be on speaking terms with Karin or she will be reluctant to tell Karin
anything.
\end{itemize}

Preconditions also act as constraints on story values. For instance, Lovisa's love for Niklas can be expressed as an interval $(0\dots9)$. Let us call an update of this interval a parameter update, and let us define a \emph{radical parameter update} as a parameter update in which the value changes with at least five units. Given that a good dramatic arc should slowly oscillate between climactic scenes and quieter ones, we can consider, e.g., oscillations between states with at least one radical parameter update and states without such updates. The oscillatory sequence of scenes can now be achieved through preconditions constraining certain transitions.

\subsection{The Set of Effectors $E$}
\label{sect:theeffectorse}
%Finally, the system maintains a database that contains information about the story world, e.g., which objects are available, the time of day, and which agents are present. Note that some information such as what an agent knows, the relationship between two agents, or even agent goals could either be stored in the database or individually with each agent. However, since it makes no difference for our purposes, and makes our discussion easier to follow, we will assume that all information about the world including information owned by any agent is stored in a global database. The data need not be restricted to first-order facts, but could include meta-information.
In the following discussion we will, for simplicity, assume that all system parameters including agent goals, plans, intentions and relationships are stored in a global database.

An effector will in Kaktus be a function updating a parameter value in the database. We thus consider parameter updates to be atomic operations on $S$, or the environmental updates to $S$ (cf. \cite{ro74}, p.248). In order to drag the plot from an undesired to a desired state, a single parameter update will hardly ever suffice. Depending on how much $S$ needs to be altered any number of updates may be necessary. We amalgamate our terminology with that of McKee to achieve the following list of value-changing actions, in increasing order of importance to the plot: 
\begin{itemize}
\item parameter update
\item beat
\item scene
\item scene sequence
\item act
\end{itemize} 

%At least a beat is required, involving perhaps a dozen effectors. Sometimes a scene change, which might require several dozen effectors, is required. 
A fairly small subset of $E$ will typically constitute a beat. A larger subset of $E$ is required for a scene change. Scene sequences and acts pertain to aspects of story design so domain specific that they will be left out of our discussion. 

%The set of control parameters used can be grouped according to scope. At the topmost level we have input from the environment, e.g., player input. At the next level are global parameters, such as time, location of agents and objects, gravity, temperature, and amount of daylight. Local parameters affect only parts of the system, e.g., the state of an object (such as a light being on or off) or the goal of an agent. The set of effectors $E$ can be divided into classes partially determined by the scope of the control parameters they are updating:
%\begin{enumerate}
%\item Environmental, affects the input to the object system $S$
%\item Global, affects the whole of $S$ at some level
%\item Local, affects only parts of $S$ (for the time being)
%\item None, has no immediate effect on $S$ (apart from the update itself) but may have so in the future
%\end{enumerate}

Below we give a short list of possible effectors in the Kaktus scenario:
\begin{itemize}
\item Simulate a player action, i.e. pretend that the player did something she actually did not.
\item Filter out some player action.
\item Introduce/remove a character.
\item Alter the flow of time to bring about or delay an event.
\item Start/stop a topic of discussion.
\item Create a disruptive event such as throwing a temper tantrum or starting a row with another character.
\item Give the player a hint or provide some information.
\end{itemize}

%The reason why we consider the last example to belong to the \emph{none} class instead of the \emph{local} class is that there is no guarantee that the player will immediately act on the information, or indeed act at all.
Effectors can have different functions. Weyhrauch divides his so-called Moe Moves, which roughly correspond to effectors, into \emph{causers, deniers, delayers, substitutions,} and \emph{hints}. Their respective function is to cause an event, stop an event from happening or deny access to an object, delay an event, substitute an event with combinations of other events, and finally give hints about unexplored paths. %These are all complementary ways of differentiating between effectors and can be used in conjunction. 

Some effectors may not have any immediate impact on the plot e.g., turning on the light in an empty room or placing an item in that room. Such effectors can, however, create favorable conditions for alternative plot unfoldings in the long term. For instance, the player might find the item placed in the room and use it to overcome an obstacle at some later instant. 

Other effectors can have a growing influence over time. Imagine for instance an effector instructing an agent to kill every other agent it encounters. At first such an act would have limited impact on the plot but as the agent roamed the world killing other agents, the effect would become increasingly noticeable. %Finally the impact of effectors applied to the environmental inputs to $S$ may range from \emph{global} to \emph{none} in scope depending on how the input was altered. 

It is important to remember that while the revision of story values during the drama describes the intended dramatic arc, these values are never directly manipulated. For instance, Lovisa's love for Niklas is never directly increased or decreased. Instead they are updated as a result of tuning other system parameters. 

Finally, we wish to stress the importance of creating effectors that do not tweak parameters in a way that interferes with user experiences of the narrative. There should be no unexplained or unexplainable twists or turns to the story brought on by the application of any effectors. It is important that transitions from one state of affairs to another are made accessible and understandable to the player \cite{se98}. Hence the design of good effectors will likely require equal amounts of artistic work and engineering.

\subsection{The Top-Down Design Process}
\label{sect:topdowndesign}
Our top-down design process for the modeling of an interactive drama (after the initial cast of characters, rudimentary plot design, and means to player interaction have been determined) consists of the following steps.
\begin{enumerate}
\item Describe the entire scenario as a finite automaton
\item For some state/transition pairs, list the resulting state
\item Partition the class of states into desirable/undesirable states
\item Partition the class of desirable states into ordinary/end states
\item Review the graph of the automaton, and iterate from 1 if necessary
%\item Minimize the automaton
%\item Begin the (real or simulated) interactive drama in real-time
%\item Maintain parameter database
\end{enumerate}

We will now review these steps in turn, in order to further explain $M$ and $E$. The first step is ideally carried out by a team of authors. The team lists a number of scenes, and then for each scene, a number of beats. The scenes must then be linked into sequences by means of scene transitions. For each scene transition, a number of parameters and their required values are identified and arranged in a database. The first output of the team may be rudimentary, approximately at the level of our Kaktus scenario. The artistic part of the work carried out should affect the implementation, and vice versa, to some extent. For instance, detailing the parameters listed, such as when pondering whether life/death is a Boolean parameter, or an interval, has an artistic aspect as well as a direct influence on the implementation in the database of parameters and their values. %We wish to stress the importance of creating effectors that do not tweak parameters in a way that interferes with user experiences of the narrative. There should be no unexplained or unexplainable twists or turns to the story brought on by the application of any effectors. It is important that transitions from one state of affairs to another are made accessible and understandable to the player \cite{se98}. Hence the design of effectors will likely require equal amounts of artistic work and engineering. 

The really hard work begins in step 2. The reason step 2 does not read ``For each state/transition pair, list the resulting state'' is that this task is insurmountable and also unnecessary. It is insurmountable because the number of transitions is equal to the number of combinations of parameter values in the entire database, which for interesting portions of dramas will be huge. It is unnecessary because most of these combinations will not, and sometimes even cannot, occur. In fact, the objective of our anticipatory system is to steer clear of more than 99 per cent of these combinations. We will therefore happily leave many transitions non-specified, and seemingly also the execution of $M$ undetermined. We will show why this is not problematic. 

\begin{figure}
\centerline{\psfig{figure=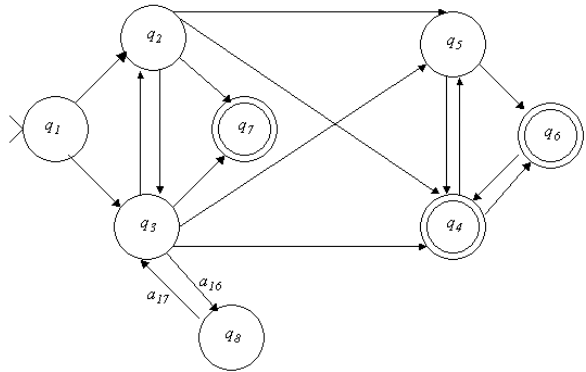,height=5.2cm}}
\caption{A state diagram for the finite automaton $M$, with one undesirable state.}
\label{fig:finite2}
\end{figure}

Since each state in $M$ is a scene, listing only the interesting transitions between these scenes is likely to result in an automaton in which all states are desirable. Hence, much of the work in step 3 is already done. Steps 2 and 3 also reflect artistic vision and imagination, however, in that the authors should try to imagine interesting undesirable states too. Figure~\ref{fig:finite2} shows a transition $a_{16}$ leading to a situation in which Karin (alias the player) expresses an interest in Niklas. This is clearly an undesired state since we have no scene dealing with such a situation. Furthermore it prevents us from using scene $q_3$ where Lovisa's love for Niklas is revealed. However it is possible to recover from this situation, e.g., by revealing that Karin's declaration of love was only a joke. Here, transition $a_{17}$ takes us back to desired territory. The necessity of explicitly listing undesirable states, and not simply stipulating that the listed states are desirable and the rest undesirable, is explained by the fact that the anticipatory system must be resilient. This resilience is a form of fault tolerance: if the changes to the interactive drama that $M$ suggests requires too radical parameter updates to be performed within one scene, or if there is a latency in the real-time parameter updating, the drama may be in limbo for a limited time. This means that $M$ will be in an undesirable state during that time. Note that analogously to the constraints placed on transitions from one climactic scene to another we can place constraints on the number and magnitude of parameter updates allowed by a single scene transition. If an undesirable state and its transitions leading back to a desirable state can be specified in advance, the drama can be led back on a desirable path. The player might in this case feel that the plot goes weird on her temporarily, but since things go back to normal, or at least come closer to what is expected by the player, after a short time, she might accept that she could not fully understand what was going on. The tuning of resilience procedures is a delicate matter and must be tested empirically. Since we do not explicitly model the player in the way Weyhrauch \cite{we97} and others do, we envisage such tests to be user studies. We recognize that there might be tacit iteration between steps 2 and 3, but their identification here as discrete steps is motivated by the fact that the number of such iterations will decrease with increased experience of the suggested design process.

Step 4 is relatively simple. Even if the preceding steps were only moderately successful, step 4 is meaningful already for automata with only desirable states; that $M$ is in a desirable state does not entail that the drama could stop in any state. Instead, most desirable states will have transitions to one or more end states, in which the plot comes to a natural conclusion. If we for instance consider interactive computer games, $M$ will normally have a single end state, in which the end credits are rolled out and all game processes are then killed. Note that an undesirable state cannot be an end state.

Step 5 is a return to the discussions that largely ran in step 1. The depiction of $M$ as a state diagram is a pedagogical and instructive basis for self-criticism and revision. Since this step depends so largely on the form of interactive drama and on the working group, we will not pursue the reasons for iteration here.

\subsection{Interaction between $M$ and $S$}
In order to execute $M$ as a string acceptor as efficiently as possible, its design must be converted to canonical form. From the Myhill-Nerode theorem follows that each automaton can be minimized with respect to its number of states (see, e.g., \cite{houl69}). There are several well-known efficient minimization algorithms, with the perhaps most widely used being Hopcroft's algorithm \cite{ho76} which is included, e.g., in the PetC automata emulation tool (freely available at \url{http://www.dsv.su.se/~henrikbe/petc/petc.html}), developed by Henrik Bergstr\"om \cite{be98}. Hopcroft's algorithm requires as input a deterministic automaton. Hopcroft's algorithm runs in $\bigcirc$($n$ log $n$), but the computational complexity is not that important to us, since it is carried out in batch. In PetC, an algorithm due to Brzozowski \cite{br62} that can minimize non-deterministic automata is included. Brzozowski's algorithm has exponential worst-case complexity but is in practice often faster than Hopcroft's algorithm \cite{wa95}. However, even small automata may experience its exponential nature \cite{glga97}. 

Although there is no difference with respect to expressive power between the classes of non-deterministic automata and their subclass of deterministic automata, i.e. they accept the same class of languages, we must pinpoint which kind of non-determinism we allow for. There are three reasons for non-determinism in an automaton in general:
\begin{itemize}
\item Transitions are labeled with strings of length different from 1
\item Undefined transitions
\item Choices between transitions with identical labels
\end{itemize}

As explained earlier, our transitions can be labeled by fairly long strings, since each transition may have to fulfill a large number of preconditions in terms of plot parameter values. However, a string label $a_1a_3a_4$ from $q_3$ to $q_4$ for example, is easily converted to a sequence of transitions: from $q_3$ to a new state $q_3'$ labeled $a_1$, from $q_3'$ to another new state $q_3''$ labeled $a_3$, and from $q_3''$ to $q_4$ labeled $a_4$. Note that the designer would never see the two new states, as we are now describing a kind of compilation procedure, so the objection that $q3'$ and $q3''$ do not correspond to scenes is irrelevant. For completeness, we should also stress that we cannot allow for labels of length 0, i.e. the empty string. This is of no significance to our construction of $M$. 
Just like our string labels can be seen as shorthand, so can our incompleteness with respect to the state/transition pairs mentioned in step 2, which leads to undefined transitions. We simply stipulate that each transition not depicted in the state diagram of $M$ leads to a dead state, i.e. a state with no edges directed to another state in $M$. This dead state is our garbage bin, and is naturally an undesired state. The only reason for non-determinism out of the three in the above list that we would like to recognize as significant is choice. We would like to allow for one out of several different scenes to follow a given scene stochastically. In an interactive drama with this possibility, the player may act in an identical fashion in two separate game rounds and still experience different unfoldings of the plot. This type of non-determinism is thus not only tolerated, but encouraged, in order for the drama to become less predictive.

So, at long last the drama begins. The appropriateness of simulated player versus real user tests depends on the maturity of the narrative. A special case is drama without human agents. In the early stages of development, this can provide ideas for plot, e.g., through plot emergence. At the other end of the spectrum is the special case of only human agents. If no artificial agents are involved in the drama, only the environment will be manipulated. As the number of human agents increases, so does in general the difficulty of maintaining a coherent plot, even though conversation between human agents might turn into an essential part of the drama, and so assume the role of a carrier of the plot.
The instant the drama begins, so does the execution of $M$. The emulation of actions in the drama is coded in the parameter database and in the state of $M$. The automaton will monitor changes made to the parameter database via the operators implemented in the drama. In the other direction, the implemented agents will experience the effects of changes instigated both from their own actions and as a result of the look-ahead in $M$: for each detection of the possible transition to an undesirable state, $M$ will make parameter updates so as to reduce the risk of, or eliminate, this possibility.

\section{Conclusions and Future Research}
\label{sect:concl}
We have described ongoing work on providing authors of interactive narratives with tools for subtle guidance of plot. The anticipatory system used for look-ahead is under implementation, and will rest in part on earlier computer science efforts to anticipatory learning \cite{da96,bodakuve00}. 

Parts of our implemented system are being developed within MagiCster, an EC-funded project that explores the use of animated, conversational interface agents in various settings. Aspects of this work addressed within MagiCster include beat selection (or dialog management), a topic not covered in depth above. We have also omitted the demonstration of the efficiency of executing $M$. While it is well-known that the computational complexity of emulating execution of small automata is low, we still need to secure its faster-than-real-time properties. This will be done in connection with the implementation of our new version of the Kaktus game.

\section*{Acknowledgments}
The authors would like to thank Paul Davidsson for comments. Laaksolahti was financed by the MagiCster and INTAGE projects, while the VINNOVA project TAP on accessible autonomous software provided Boman with the time required for this study.

\bibliographystyle{plain}
\bibliography{bookfinal}

\end{document}